\documentclass{article}
\usepackage{main}

\usepackage{times}
\usepackage{multirow}
\usepackage{colortbl}
\usepackage{soul}
\usepackage{url}
\usepackage[hidelinks]{hyperref}
\usepackage[utf8]{inputenc}
\usepackage[small]{caption}
\usepackage{graphicx}
\usepackage{amsmath}
\usepackage{amssymb}
\usepackage{xcolor} 
\usepackage{amsthm}
\usepackage{booktabs}
\usepackage{algorithm}
\usepackage{algorithmic}
\usepackage[switch]{lineno}

\title{MC-VTON: Minimal Control Virtual Try-On Diffusion Transformer}

\author{
Junsheng Luan\textsuperscript{*}$^1$
\and
Guangyuan Li\textsuperscript{*}$^1$\and
Lei Zhao$^1$\And
Wei Xing $^1$\\
\affiliations
$^1$Zhejiang University\\
\emails
\{l.junsheng121, cslgy, cszhl, wxing\}@zju.edu.cn
}

\begin{document}

\twocolumn[{
    \renewcommand\twocolumn[1][]{#1}
\maketitle

\begin{center}
        \captionsetup{type=figure}
        \includegraphics[width=0.95\linewidth]{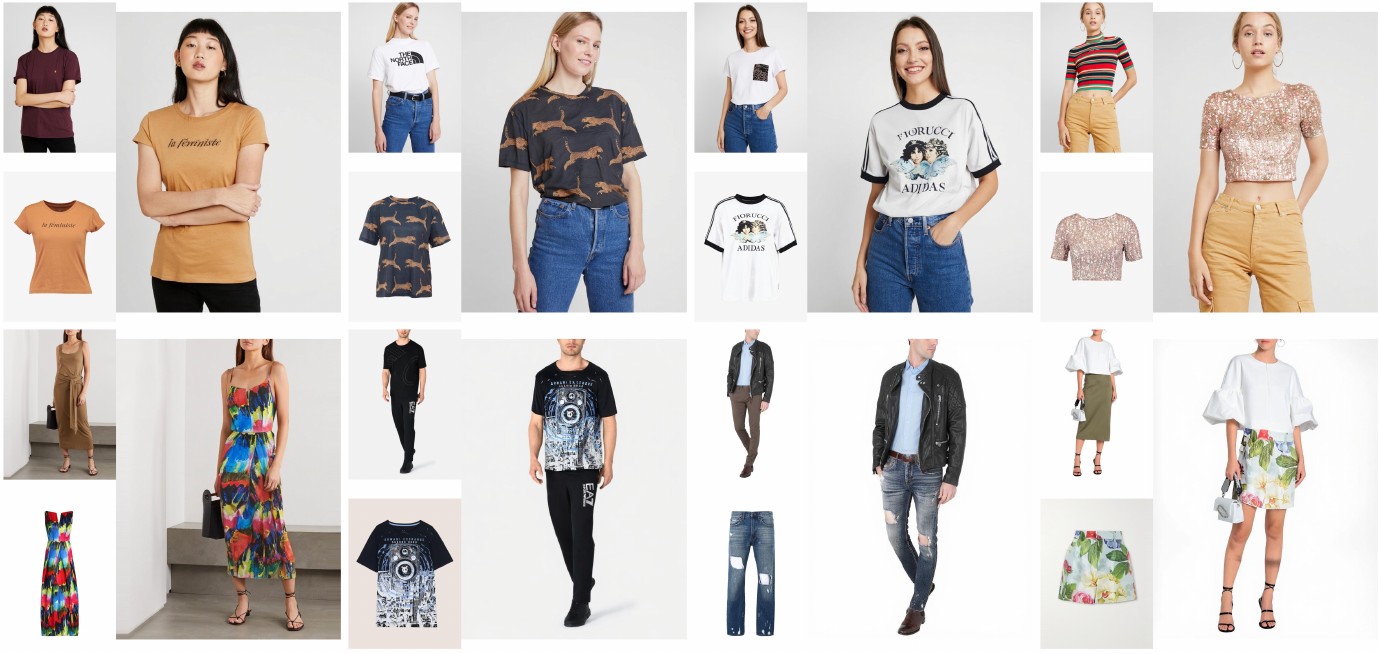}
        \captionof{figure}{MC-VTON achieves superior performance on virtual try-on tasks with fine-grained detail preservation, simplified network and inputs, less training parameters, and less inference steps.}
        \label{head}
\end{center}
}]

\renewcommand{\thefootnote}{}
\footnotetext{*Equal contribution.}
\begin{abstract}
Virtual try-on methods based on diffusion models achieve realistic try-on effects. They use an extra reference network or an additional image encoder to process multiple conditional image inputs, which adds complexity pre-processing and additional computational costs. Besides, they require more than 25 inference steps, bringing longer inference time. In this work, with the development of diffusion transformer~(DiT), we rethink the necessity of additional reference network or image encoder and introduce MC-VTON, which leverages DiT's intrinsic backbone to seamlessly integrate minimal conditional try-on inputs. Compared to existing methods, the superiority of MC-VTON is demonstrated in four aspects: (1)~\textbf{Superior detail fidelity.} Our DiT-based MC-VTON exhibits superior fidelity in preserving fine-grained details. (2)~\textbf{Simplified network and inputs.} We remove any extra reference network or image encoder. We also remove unnecessary conditions like the long prompt, pose estimation, human parsing, and depth map. We require only the masked person image and the garment image. (3)~\textbf{Parameter-efficient training.} To process the try-on task, we fine-tune the FLUX.1-dev with only 39.7M additional parameters~($0.33\%$ of the backbone parameters). (4)~\textbf{Less inference steps.} We apply distillation diffusion on MC-VTON and only need 8 steps to generate a realistic try-on image, with only 86.8M additional parameters~($0.72\%$ of the backbone parameters). Experiments show that MC-VTON achieves superior qualitative and quantitative results with fewer condition inputs, trainable parameters, and inference steps than baseline methods. 
\end{abstract}

\section{Introduction}
Virtual try-on (VTON) is a technology that overlays specific clothing onto user photos. It has aroused widespread attention for its potential applications in the e-commerce industry like online shopping. In recent years, many researchers have made significant efforts toward achieving more realistic and photorealistic virtual try-on results based on diffusion models. Most of them adopt an Encoder-Backbone structure, which consists of a garment encoder and a denoising backbone U-Net. The garment encoder can be a CLIP image encoder~\cite{ip-adapter}, a ControlNet~\cite{controlnet} or simply replicates the backbone network, which encodes the garment feature. Backbone U-Net processes the multiple conditional image inputs and extracts person features, then utilizes attention mechanisms to facilitate interaction between the garment and person features.

Although these methods achieve realistic try-on effects, they have the following drawbacks: (1) Missing garment details. The texture on the given garment image is not generated correctly. (2) Complex network and inputs. For network, most methods apply an additional reference network or image encoder. For inputs, IDM-VTON, StableVTON and TryOnDiffusion use extra DensePose input to extract fine-grained human pose feature; TryOnDiffusion also takes garment pose as extra inputs; FitDiT uses extra DWPose input to extract pose feature to guide the DiT denoising process. Some also use long prompts to guide the generation. These condition inputs bring complex pre-processing and a huge computational burden. (3) Substantial training parameters. The extra network of these methods brings a substantial number of training parameters. (4) High inference steps. These methods usually require more than 25 inference steps to generate a try-on image, bringing longer inference time. 

In this work, with the development of diffusion transformer~(DiT)~\cite{dit,fitdit,sd3,flux}, we rethink the necessity of reference network and propose MC-VTON, which leverages DiT’s intrinsic backbone to seamlessly integrate minimal conditional try-on inputs. To address the first drawback, we leverage the strengths of the diffusion transformer architecture, focusing more attention on high-resolution latent features related to garment details, and then achieve superior fidelity in preserving fine-grained detail. For drawbacks~(2)(3), we eliminate any additional reference network or image encoder and propose a parameter-efficient approach for incorporating garment/person features into DiT architectures. The \textbf{only} conditional image inputs are the garment image and the masked person image. We reuse the VAE encoder to process them, then augment the features with learnable position embeddings, and seamlessly integrate them alongside latent noise in the denoising net. This approach enables efficient multi-modal attention interaction between the garment/person feature, which brings about only 39.7M additional parameters~($0.33\%$ of the FLUX.1-dev parameters). Finally, to address drawback~(4), we fix the trained MC-VTON and add multi heads to every transformer layer, as the discriminator backbone, then apply distillation diffusion on MC-VTON, with 86.8M additional parameters~($0.72\%$ of the FLUX.1-dev parameters). Our contributions are summarized as follows:
\begin{figure}[t!]

    \includegraphics[width=0.48\textwidth]{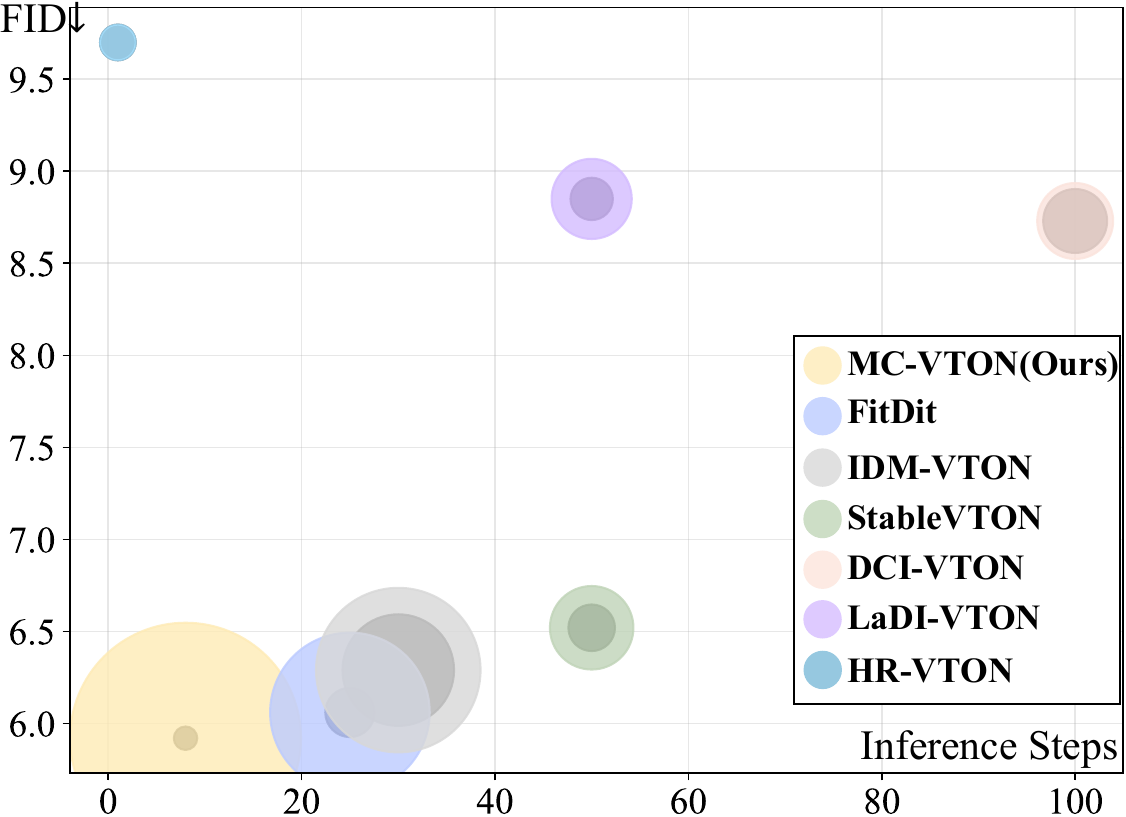} 
    \caption{Efficiency comparison with try-on methods. Each method is represented by two concentric circles. The outer circle denotes the total parameters and the inner circle denotes the trainable parameters. MC-VTON achieves lower FID on the VITONHD dataset with fewer trainable parameters and inference steps.}
    \label{headcomp}
\end{figure}
\begin{itemize}
    \item We propose a DiT-based virtual try-on model MC-VTON, leveraging the strengths of the diffusion transformer architecture, focusing more attention on high-resolution latent features related to garment details, exhibiting superior fidelity in preserving fine-grained details.
    \item We eliminate any additional reference network or image encoder, and propose a parameter-efficient approach for incorporating garment/person features into DiT architectures. This enables DiT to integrate only two conditional image inputs by utilizing its intrinsic backbone, introducing minimal training parameters.
    \item By utilizing distillation diffusion, we effectively reduce the number of inference steps from over 25 to just 8.
    \item Extensive qualitative and quantitative experiments have demonstrated MC-VTON's superiority over state-of-the-art virtual try-on models. It achieves competitive inference time of 5.23 seconds for a single $1024 \times 768$ image and utilizes the minimal training parameters of 136.5M, paving the way for more sophisticated real-world applications.

\end{itemize}

\section{Related Work}

\subsection{Conditional diffusion generation}
Text-to-image diffusion models~\cite{ddpm,LDM,GLIDE,ddim,Cascade} have shown the capability of generating high-quality images with a single prompt. However, complex image synthesis tasks like virtual try-on require fine-grained control. To this end, recent researches add conditional image control to T2I diffusion models with extra networks. For instance, ControlNet~\cite{controlnet} and T2I-Adapter~\cite{T2I} propose to fine-tune an extra reference network that encodes spatial information such as edges, depth, and human pose, to control the diffusion model together with text prompts. IP-Adapter~\cite{ip-adapter} proposes to use an extra image encoder to extract high-level semantic features of reference images, and control image generation with both textual and visual prompts. These methods bring extra network modules and substantial training parameters. In this paper, we utilize the intrinsic backbone of diffusion transformer to process the conditional image inputs, instead of introducing extra modules and substantial training parameters.

\subsection{Image-based Virtual Try-On}
Image-based virtual try-on has been extensively researched over the years, emerging as a promising task. Image-based virtual try-on aims to synthesize an image of a person who wears a specified garment while maintaining identity and consistency. There have been warping-based and diffusion-based methods. Generally, warping-based methods decompose the task into training a warping module and an image generation module, where the image generation is based on the warping result. For instance, VITON~\cite{viton} utilizes a coarse-to-fine TPS transformation pipeline for garment warping. CP-VTON~\cite{cpvton} explicitly divides the try-on strategy into the warping and generation stages. GP-VTON~\cite{gpvton} uses local flow to warp garment components and global parsing to compose them, improving detail consistency and visual quality. However, they have limitations in that they fail to align well between garment and person, bringing visual distortions. 

With the development of text-to-image diffusion models~\cite{LDM}, diffusion-based methods achieve realistic try-on effects and solve the above limitations. Specifically, for U-Net architecture, LaDI-VTON~\cite{ladi-vton} utilizes textual inversion within the diffusion model to map the visual features of garments into the CLIP embedding space. DCI-VTON~\cite{DCI-VTON} uses the warped garment as a conditional input to the diffusion model to better preserve the garment’s features. IDM-VTON~\cite{IDM-VTON} utilizes two different modules to encode the semantics of garment images. For DiT architecture, FitDiT~\cite{fitdit} proposes the first DiT-based try-on model and exhibits high fidelity in preserving fine-grained details. However, these methods use an extra reference network or an additional image encoder to process multiple conditional image inputs, bringing complex inputs and substantial training parameters. Our proposed MC-VTON eliminates any additional reference network or image encoder. It integrates two conditional image inputs using itself while introducing a minimal increase in training parameters.

\begin{figure*}[t!]
    \centering
    \includegraphics[width=1\textwidth]{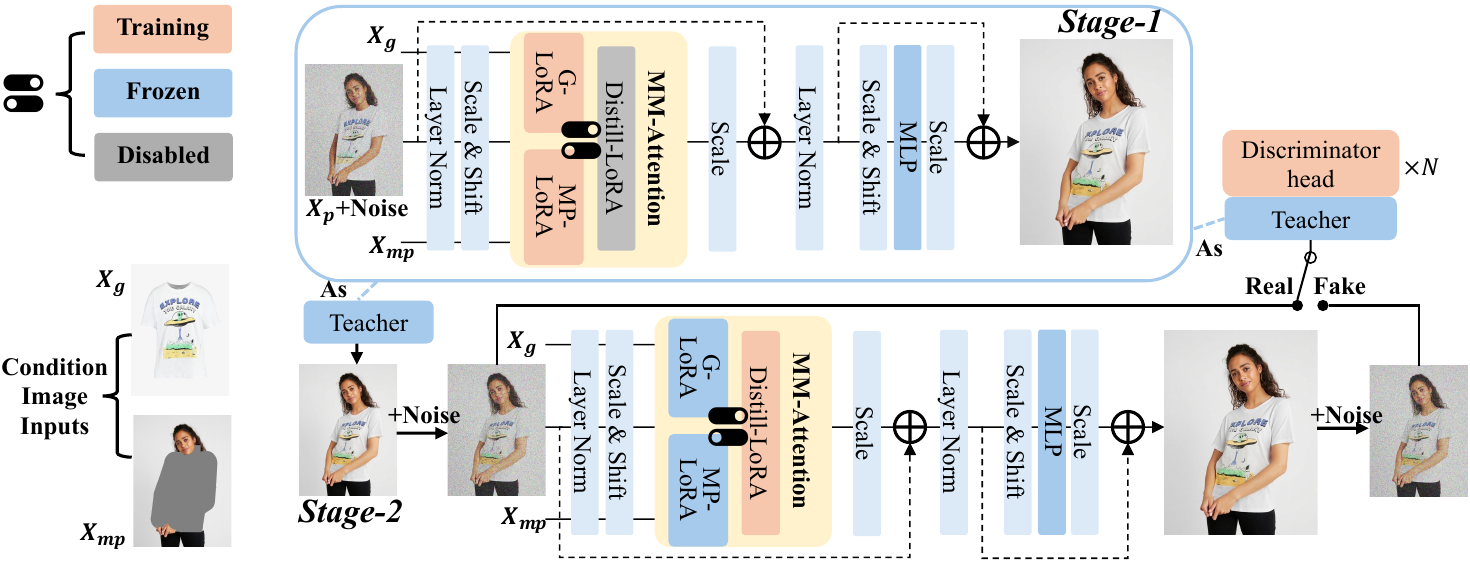} 
    \caption{The architecture of our proposed MC-VTON. Our model is trained in two stages. The upper-left LoRA-switch controls the statuses of three types of LoRA: G-LoRA, MP-LoRA, and Distill-LoRA. Stage-1 trains the try-on ability. G-LoRA and MP-LoRA are trained and Distill-LoRA is disabled. Stage-2 applies distillation diffusion. The G-LoRA and MP-LoRA are frozen. The trained MC-VTON acts as the teacher when Distill-LoRA is disabled, and acts as the student when Distill-LoRA is training.}
    \label{workflow}
\end{figure*}

\section{Methods}

\subsection{Preliminary}

Our method is built on top of diffusion transformer~(DiT)~\cite{dit}, which is employed in models like FLUX.1~\cite{flux}, Stable Diffusion 3~\cite{sd3}. It uses a denoising network of transformer blocks to denoise noisy image tokens iteratively. In a transformer block, the inputs are image tokens and text condition tokens, which are first embedded into image embedding $X$ and text embedding $C_T$. Then, they are passed into the AdaLN-Zero~\cite{dit} layer, where the embeddings are normalized, scaled, and shifted as follows:
\begin{equation}
X^* = \alpha_x \cdot \hat{X} + \beta_x ; ~~ C_T^* = \alpha_{ct} \cdot \hat{C_T} + \beta_{ct} 
\end{equation}
where $\hat{X}$ and $\hat{C_T}$ are the normalized embeddings. $\alpha$ and $\beta$ are corresponding scale parameters and shift parameters, which are generated with the text and timestep embeddings using an MLP. FLUX.1 proposes rotary position embedding~(RoPE) to incorporate positional dependencies across tokens. Specifically, for token in position $(i,j)$, $X^*_{i,j}$ is multiplied with $R_{i,j}$ where $R$ is the rotation matrix. This captures detailed associations between local areas of the image, improving model generation quality. $C_T^*$ is processed similarly but with the corresponding rotation matrix set to zero. The embeddings of $X^*$ and $C_T^*$ are then projected to query, key embeddings; value embeddings are projected from $X^*$ and $C_T^*$ without RoPE operation.

Finally, their embeddings are concatenated with a unified sequence $Z = [X; C_T]$, generating $Q_Z$, $K_Z$ and $V_Z$. They are applied MM-Attention calculation as 
\begin{equation}
Attention(Z) = softmax(\frac{Q_Z K^{\top}_Z}{\sqrt{d}}V_Z)
\end{equation}
which facilitates interaction between the text and image features. Above are the primary FLUX.1 generation without conditional image inputs. In the following sections, we integrate two conditional image inputs based on FLUX.1.

\subsection{MC-VTON}

\subsubsection{Model Overview}

Our model is trained in two stages. Stage-1 trains the try-on ability and Stage-2 trains the accelerated generation. An overall framework of MC-VTON is presented in Fig.\ref{workflow}. The \textbf{only} conditional image inputs are the garment image $X_g$ and the masked person image $X_{mp}$. Our goal is to utilize DiT's intrinsic backbone to integrate the two conditional image inputs without any extra reference network or image encoder.

During training, we first encode the ground truth person image $I_{p}$ with DiT VAE and add noise to it. We then use VAE $\mathcal{E}$ to encode $I_{g}$ and $I_{mp}$ and get the conditional image tokens $X_g$ and $X_{mp}$. These image tokens are altogether input into the transformer blocks. For text condition tokens, we fix the text to a single word "Model" instead of long descriptions of the garment and the person, which proves to be effective enough in practice.

\subsubsection{Integrating Try-On Features}
It is common practice to treat virtual try-on as a specific case of subject-driven image inpainting task: $X_{g}$ for \textit{subject-driven generation task} and $X_{mp}$ for \textit{image inpainting task}. We fill the masked region in $X_{mp}$ with the visual features of $X_{g}$, like the shape, color, and text of the pattern on the garment. The text encoder encodes the word ``Model'' to text condition tokens $C$, then with the conditional image tokens are integrated into a unified sequence: $[X_p;X_g; X_{mp};C]$, and is then embedded, normalized, scaled and shifted, then passed into MM-Attention block. This enables direct interaction in multi-modal attention without additional reference networks. MM-Attention block processes the try-on task by seamlessly integrating $X_g$ and $X_{mp}$ alongside $X_p$, facilitating interaction between the garment/person feature. 

\subsubsection{Advanced RoPE}
Since FLUX.1 proposes RoPE to incorporate positional dependencies across tokens, we need to perform them on conditional image tokens $X_g$ and $X_{mp}$ to ensure effective interaction with the noisy image tokens $X_p$. For a $768\times1024$ garment/person image, the VAE encodes it and divides it into $48\times64$ grid of tokens, then each token is assigned a unique two-dimensional position index $(i, j)$ with $i \in [0, 47]$ and $j \in [0,63]$. Following ~\cite{ominicontrol}, for the spatially aligned task, image inpainting, we keep the position index to the original settings in FLUX.1; for the non-spatially aligned task, subject-driven generation, we shift the position index $j$ to $[64,128]$ to avoid spatial overlap. This leads to faster convergence during training. The Stage-1 training uses the following denoising loss function:
\begin{equation}
L = \mathbb{E}_{\epsilon \sim \mathcal{N}(0,1), t \sim \mathcal{U}(t)} \left[ w(t) \left\| \epsilon_\theta(X_t, [X_g; X_{mp}; C]) - \epsilon \right\|^2 \right]
\end{equation}
where $X_t = (1 - t)X_p + t \epsilon$ is the noisy image tokens~\cite{rectifiedflow}, and $w(t)$ is a weighting function at each timestep $t$.

\subsection{Parameter-Efficient Joint Training}

Since MC-VTON utilizes DiT's intrinsic backbone to integrate the two conditional image inputs, we can simply fine-tune the FLUX.1-dev model with denoising loss. To avoid fine-tuning the huge model with 12B parameters, we perform LoRA~\cite{lora} fine-tuning which significantly reduces the time and memory consumption.

However, the try-on task indeed composes the image inpainting task~(spatial aligned) and the subject-driven generation task~(non-spatial aligned). The added LoRA meets conflict during the joint training of the two tasks. This slows down the convergence and brings distortion to the generation. We thus propose G-LoRA, MP-LoRA to deal with the subject-driven generation task and the image inpainting task, respectively. We further propose LoRA-switches to better control the LoRAs. As shown in Fig.~\ref{workflow}, a LoRA block has three statuses: disabled, frozen, and training. The embedding will not be affected by a disabled LoRA~(by setting the LoRA scale to zero); a frozen LoRA affects the embedding but the parameters are not optimized. 

We add G-LoRAs, MP-LoRAs, and the LoRA-switches to the QKV Linear layers in the MM-Attention blocks. For the task separation, $X_g$ is projected to QKV embeddings with MP-LoRA disabled; $X_{mp}$ is projected with G-LoRA disabled; $X_p$ is projected with both LoRA disabled, indicating the use of the vanilla FLUX.1 image generation capability. The parameter-efficient joint training brings only 26.5M parameters from G-LoRAs and 13.2M parameters from MP-LoRAs, a total of $0.33\%$ of the FLUX.1-dev backbone. 

\subsection{Leveraging Distillation Diffusion}

\begin{figure*}[t!]
    \centering
    \includegraphics[width=0.96\textwidth]{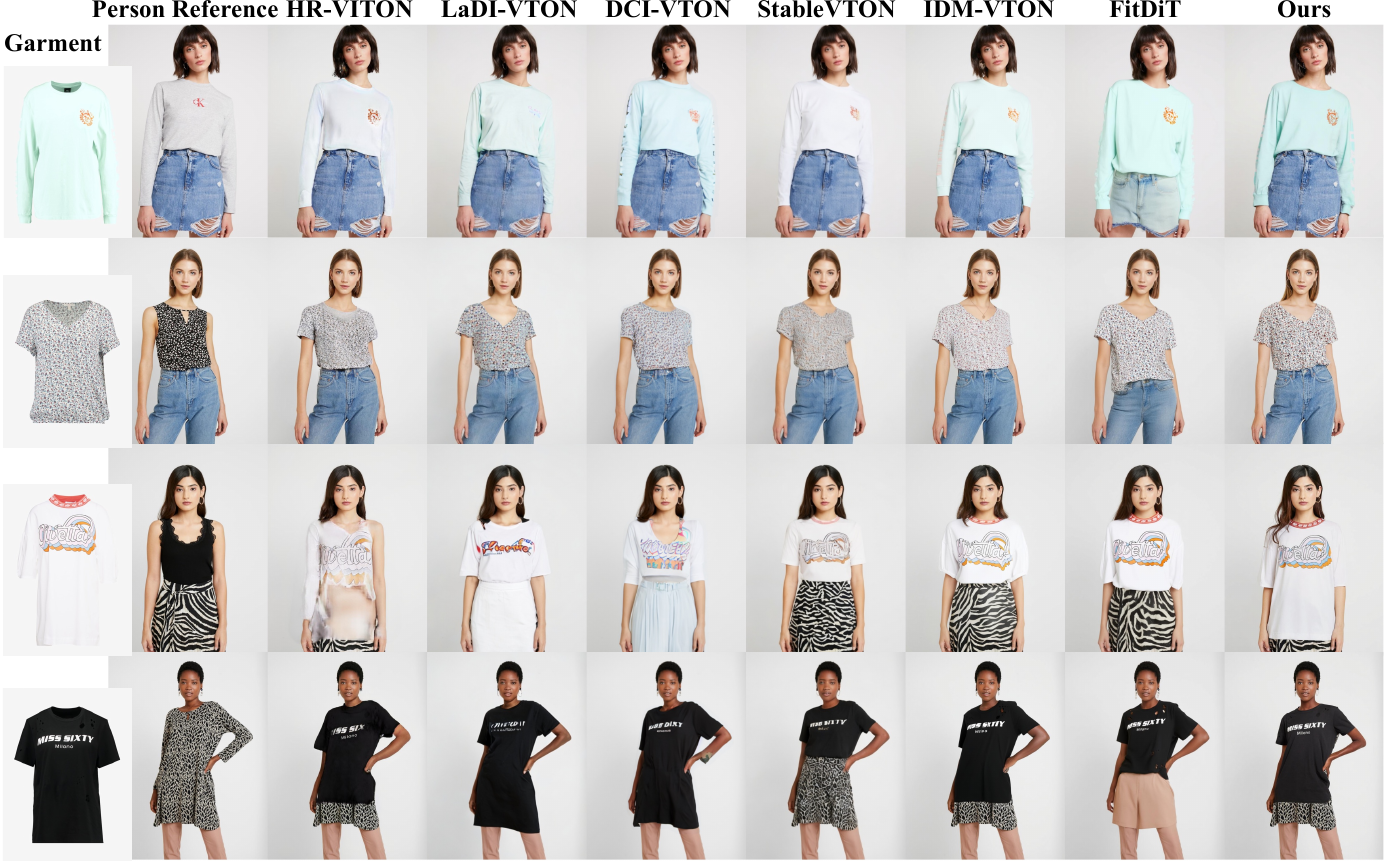} 
    \caption{Qualitative comparison of our proposed MC-VTON with other methods on VITON-HD dataset. Best viewed when zoomed in.}
    \label{qua1}
\end{figure*}

Finally, we propose to reduce the inference steps in the try-on tasks. For text-to-image generation, the focus is on the final distribution at $t = 0$ , which can be achieved through simpler differential equations with straighter trajectories, allowing larger step sizes and fewer evaluations. Following LADD, we leverage adversarial distillation diffusion in the latent space as the Stage-2 training. For parameter-efficient training, we add Distill-LoRAs to the MM-Attention blocks. We disable them in the Stage-1 training and train them in Stage-2 while G-LoRAs and mp-LoRAs are frozen.

Specifically, with a group of conditional inputs $[X_g;X_{mp};C]$, we generate a try-on image $X_0$ within 50 steps~(default setting of FLUX.1-dev) using the teacher MC-VTON~(with distill-LoRAs disabled), then add noise to it. Following \cite{sd3turbo}, the 8-step student MC-VTON~(with distill-LoRAs enabled and training) predicts $\hat{X_0}$ and renoises it with the same scale noise. We then apply the teacher model to the noised $\hat{X_0}$ and apply independent discriminator heads. The objective of the distillation is the same with ADD~\cite{add}:
\begin{equation}
\mathcal{L}_{\mathrm{adv}}^{\mathrm{G}}(\phi)  =-\sum\mathcal{D}_{\phi,k}(\hat{X_0},[X_g;X_{mp};C])
\end{equation}
\begin{equation}
\begin{aligned}&\mathcal{L}_{\mathrm{adv}}^{\mathrm{D}}(\phi) =\sum_k\max(0,1-\mathcal{D}_{\phi,k}(X_0))+\gamma R1(\phi)\\&+\sum_k\max(0,1+\mathcal{D}_{\phi,k}(\hat{X_0},[X_g;X_{mp};C])\end{aligned}
\end{equation}
where $\phi$ is the Distill-LoRA parameters, $\mathcal{D}$ is the teacher model with discriminator heads, $k$ is the head number and $R1$ denotes the R1 gradient penalty. The student model and teacher model share the same backbone, differing only in the state of the Distill-LoRA; meanwhile, the distillation process is in latent space, which does not require decoding to RGB space. This design minimizes the computational costs of Stage-2.

\section{Experiments}
\subsection{Experimental Setup}
\subsubsection{Datasets}
We employ two publicly available datasets, VITON-HD~\cite{vitonhd} and DressCode~\cite{dresscode}, to evaluate the virtual try-on task. Both datasets consist of paired images of garments and their corresponding human models wearing the garments. The DressCode dataset is categorized into three classes: upper body, lower body, and dresses. The testing experiments are conducted under two settings: paired and unpaired. In the paired setting, the input garment image and the garment worn by the human model are the same item. Conversely, the human model tries on different garments in the unpaired setting. We train MC-VTON on VITON-HD and test on VITON-HD and DressCode.

\subsubsection{Baselines}
To demonstrate the superiority of our proposed method, we compare our method with a GAN-based virtual try-on method: HR-VITON, four U-Net based virtual try-on diffusion methods LaDI-VTON, DCI-VTON, StableVTON, IDM-VTON, and a DiT based method FitDiT. For a fair comparison, all generated images are in a resolution of $1024\times768$. For methods generating $512\times384$ images, we upsample their generated images.

\begin{table*}[t!]
\centering
\resizebox{0.9\linewidth}{!}{
\begin{tabular}{lccccccccccc}
\toprule
\multirow{2}{*}{Method} & \multicolumn{5}{c}{VITON-HD} & \multicolumn{5}{c}{DressCode} & \multirow{2}{*}{Step}\\
\cmidrule(lr){2-6} \cmidrule(lr){7-11}
 & LPIPS↓ & SSIM↑ & FID↓ & KID↓ & CLIP-I↑ & LPIPS↓ & SSIM↑ & FID↓ & KID↓& CLIP-I↑ \\
\midrule
\rowcolor{gray!20}
\multicolumn{12}{c}{GAN-based methods} \\
\midrule
HR-VITON  & 0.115 & 0.883 & 9.70 &1.94& 0.832 & 0.112 & 0.910 & 21.42 &2.43& 0.771 & 1\\
\midrule
\rowcolor{gray!20}
\multicolumn{12}{c}{U-Net diffusion-based methods} \\
\midrule
LaDI-VTON & 0.156 & 0.872 & 8.85 &1.62& 0.834 & 0.149 & 0.905 & 16.54 &2.07& 0.803 & 50\\
DCI-VTON & 0.166 & 0.856 & 8.73 &1.57& 0.840 & 0.162 & 0.893 & 17.63 &1.75& 0.777 &100\\
StableVTON & 0.133 & 0.885 & 6.52 &1.23& 0.871 & 0.107 & 0.910 & 14.37 &1.53& 0.866 & 50\\
IDM-VTON & 0.102 & 0.870 & 6.29 &1.20& 0.883 & \textbf{0.062} & 0.920 & 8.64 &1.22& \underline{0.904} &30\\
\midrule
\rowcolor{gray!20}
\multicolumn{12}{c}{DiT diffusion-based methods} \\
\midrule
FitDiT & \underline{0.083}  & \underline{0.895} & \underline{6.06} &\textbf{0.759}& \underline{0.900} & 0.079 & 0.913 & 7.25  & \textbf{0.831}& 0.901 &25\\
MC-VTON & \textbf{0.066} & \textbf{0.902} & \textbf{5.92}  &\underline{0.773}& \textbf{0.903} & \underline{0.076} & \textbf{0.921} & \textbf{6.97} & \underline{0.843}& \textbf{0.907} &8\\
\bottomrule
\end{tabular}
}
\caption{Quantitative results on VITON-HD and DressCode. \textbf{Bold} and \underline{underline} denote the best and the second best result, respectively.}
\label{quan1}
\end{table*}

\begin{table*}[t!]
\centering
\resizebox{0.85\linewidth}{!}{
\begin{tabular}{lcccccccc}
\toprule
\multirow{2}{*}{Methods} & \multicolumn{2}{c}{Extra Network} & \multicolumn{3}{c}{Extra Conditions} & \multicolumn{3}{c}{Parameters~(M)} \\
\cmidrule(lr){2-3} \cmidrule(lr){4-6}  \cmidrule(lr){7-9}
 & Reference & Encoder & \multicolumn{2}{c}{Pose}& Long-text & Total & Trainable & Ratio~(\%)\\
\midrule

HR-VTON  & \checkmark & \checkmark &  \multicolumn{2}{c}{DensePose}& - & 293 & 293 & 100 \\
\midrule

LaDI-VTON & \checkmark & \checkmark & \multicolumn{2}{c}{KeyPoint}& \checkmark & 1482 & 414 & 27.9  \\
DCI-VTON & \checkmark & - &  \multicolumn{2}{c}{DensePose}& - & 1330 & 943 & 70.9  \\
StableVTON & \checkmark & - & \multicolumn{2}{c}{DensePose} & - & 1608 & 501 & 31.2 \\
IDM-VTON & \checkmark & \checkmark & \multicolumn{2}{c}{DensePose}& \checkmark & 6238 & 2871 & 46.0 \\
\midrule

FitDiT & \checkmark & - &\multicolumn{2}{c}{DWPose} & - & 5870 & 561 & 9.56 \\
\rowcolor{gray!20}
MC-VTON~(Ours) & - & - & \multicolumn{2}{c}{-} &- & 12288 & 127 & 1.03\\
\bottomrule
\end{tabular}
}
\caption{Efficiency comparison. Our proposed MC-VTON requires no extra network, no extra conditions, with the least trainable parameters.}
\label{quan2}
\end{table*}

\subsubsection{Evaluation Metrics}
We conduct quantitative evaluations under both unpaired and paired settings. Specifically, in the unpaired setting, we employ the Frechet Inception Distance (FID) and Kernel Inception Distance (KID)~\cite{kid} to evaluate the realism of the generated results. In the paired setting, where ground truth is available, we use the Learned Perceptual Image Patch Similarity (LPIPS) and Structural Similarity Index Measure (SSIM) to assess the quality of the generated images. We also use the CLIP-I~\cite{CLIP} image similarity score to evaluate high-level image similarity.

\subsubsection{Implementation Details}
Our base model is FLUX.1-dev. We utilize LoRA training on MC-VTON in two stages. For the first stage, to better learn the garment feature, we set the rank of G-LoRA to 8, while the rank of MP-LoRA is set to 4. The model is trained with a batch size of 6 and an accumulation step of 1. We employ the Prodigy optimizer~\cite{prodigy} with safeguard warmup and bias correction enabled, setting the weight decay to 0.01. For the second stage, we employ the Adam Optimizer~\cite{adam} and the LoRA rank of distill-LoRA is 16. We conduct our two-stage training on four NVIDIA A100 (80GB) GPUs with image resolutions of $1024\times768$.

\subsection{Qualitative Results}
\begin{figure}[t!]
    \centering
    \includegraphics[width=0.44\textwidth]{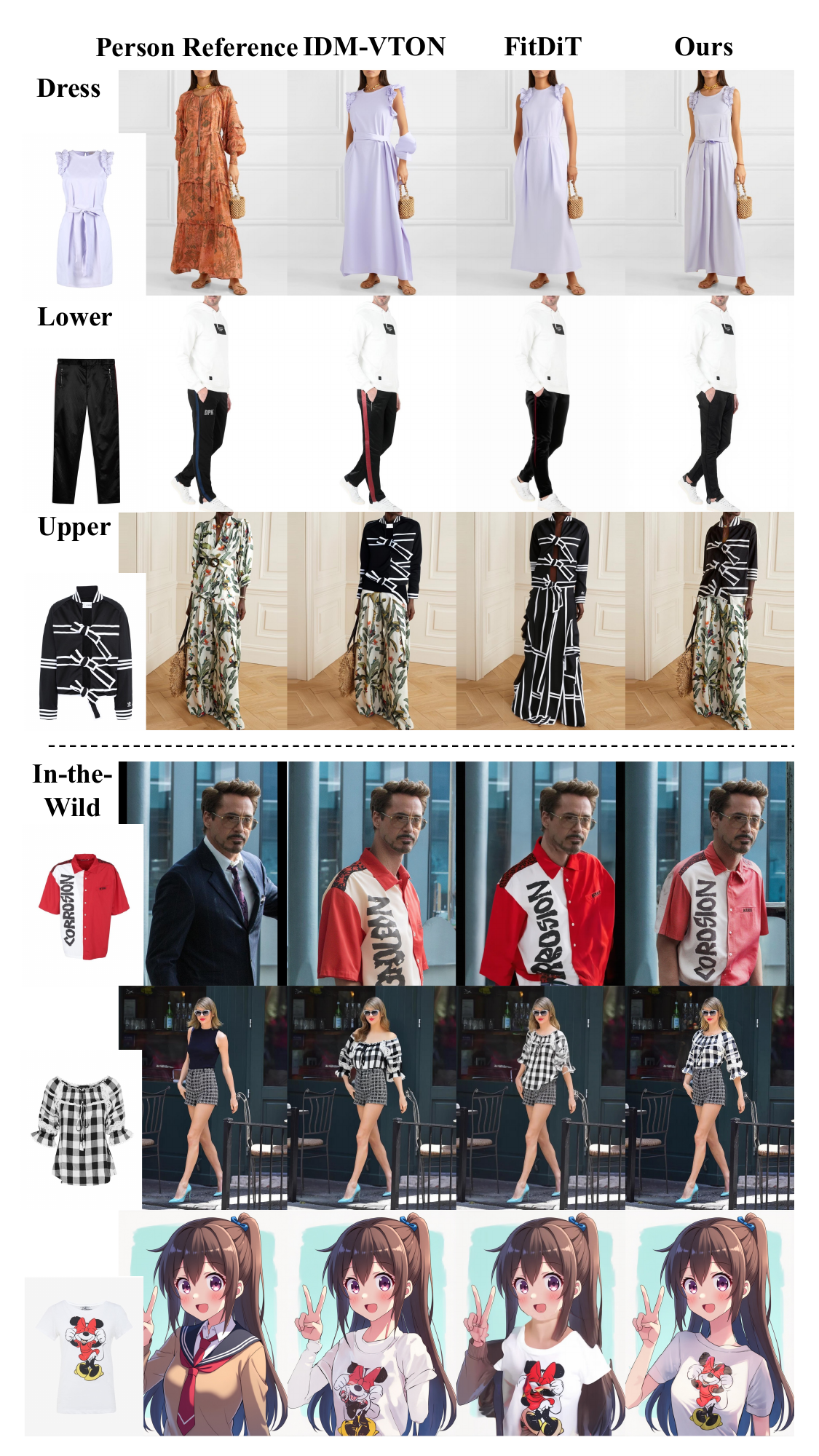} 
    \caption{Additional qualitative comparison with IDM-VTON and FitDiT. The upper group is conducted on DressCode dataset and the lower group is the in-the-wild comparison. Best viewed when zoomed in.}
    \label{qua3}
\end{figure}

We conduct a qualitative analysis under the unpaired setting of VITON-HD and DressCode. For the VITON-HD dataset, in Fig.~\ref{qua1}, MC-VTON shows superior performance in capturing fine-grained details of the garment like the text and pattern, while maintaining the consistency of the person image like the unchanged parts. We also observe that GAN-based method and U-Net based diffusion methods can not preserve detailed features well, while the DiT-based method FitDit does not keep the consistency. In Fig.~\ref{qua3}, we compare with the IDM-VTON and FiTDiT on the DressCode dataset. Additionally, we conduct virtual try-on in-the-wild, featuring complex backgrounds, human poses and style. As can be seen, our proposed MC-VTON outperforms other methods, showing a superior generalization ability and image quality.

\subsection{Quantitative Results}
Quantitative analysis is conducted under both the paired setting (evaluated using SSIM, LPIPS, and CLIP-I) and the unpaired setting (evaluated using FID and KID). As shown in Tab.~\ref{quan1}, our proposed MC-VTON outperforms other methods on most metrics within only 8 inference steps, demonstrating its effectiveness in generating high-quality and realistic try-on images. Additionally, Tab.~\ref{quan2} shows the efficiency comparison. Compared to the other methods, our proposed MC-VTON requires no extra network, no extra conditions, with the least trainable parameters.

\subsection{Ablation Study}
Since G-LoRA and MP-LoRA can not be stripped from the entire network and evaluated individually, we conduct the ablation studies on the LoRA rank of G-LoRA, MP-LoRA and Distill-LoRA. We also evaluate the effect of distillation diffusion using the VITON-HD dataset. 

\subsubsection{Ablation on LoRA Rank}
As shown in Tab.~\ref{ab1}, although a MP-LoRA rank of 16 achieves the highest FID and KID, it brings $4\times$ training parameters. Results show that a G-LoRA rank of 8, a MP-LoRA rank of 4 and a Distill-LoRA rank of 16 achieve a balance between training efficiency and try-on performance.
\begin{table}[t!]
\centering
\resizebox{0.87\linewidth}{!}{
\begin{tabular}{lccccc}
\toprule
Rank & LPIPS↓ & SSIM↑  & FID↓ & KID↓ & CLIP-I↑\\
\midrule
\rowcolor{gray!20}
\multicolumn{6}{c}{G-LoRA} \\
\midrule
2 & 0.075 & 0.893  & 6.26 & 0.850 & 0.894\\
4 & 0.073 & 0.891  &6.05 & 0.793 & 0.900\\
8~(Ours)& \textbf{0.066} & 0.902 & \textbf{5.92}  &\textbf{0.773}& 0.903 \\
16 & 0.072 & \textbf{0.903}  &6.53 & 0.851 & \textbf{0.908} \\
\midrule
\rowcolor{gray!20}
\multicolumn{6}{c}{MP-LoRA} \\
\midrule
2 & 0.080 & 0.893  & 7.87 & 0.953 & 0.900\\
4~(Ours) & \textbf{0.066} & \textbf{0.902} & 5.92  &0.773& \textbf{0.903}\\
8 & 0.069 & 0.891  &5.96 & 0.775 & 0.903\\
16 & 0.071 & 0.887  &\textbf{5.89} & \textbf{0.770} & 0.902 \\
\midrule
\rowcolor{gray!20}
\multicolumn{6}{c}{Distill-LoRA} \\
\midrule
8 & 0.093 & 0.876  & 7.63& 0.789 & 0.903\\
16 & \textbf{0.066} & \textbf{0.902} & 5.92  &\textbf{0.773}& \textbf{0.903}\\
32 & 0.068 & 0.899  &\textbf{5.85} & 0.780 & 0.900 \\

\bottomrule
\end{tabular}}
\caption{Ablation study on LoRA rank. \textbf{Bold} denotes the best.}
\label{ab1}
\end{table}

\subsubsection{Ablation on Distillation Diffusion}
\begin{figure}[h]
    \centering
    \includegraphics[width=0.44\textwidth]{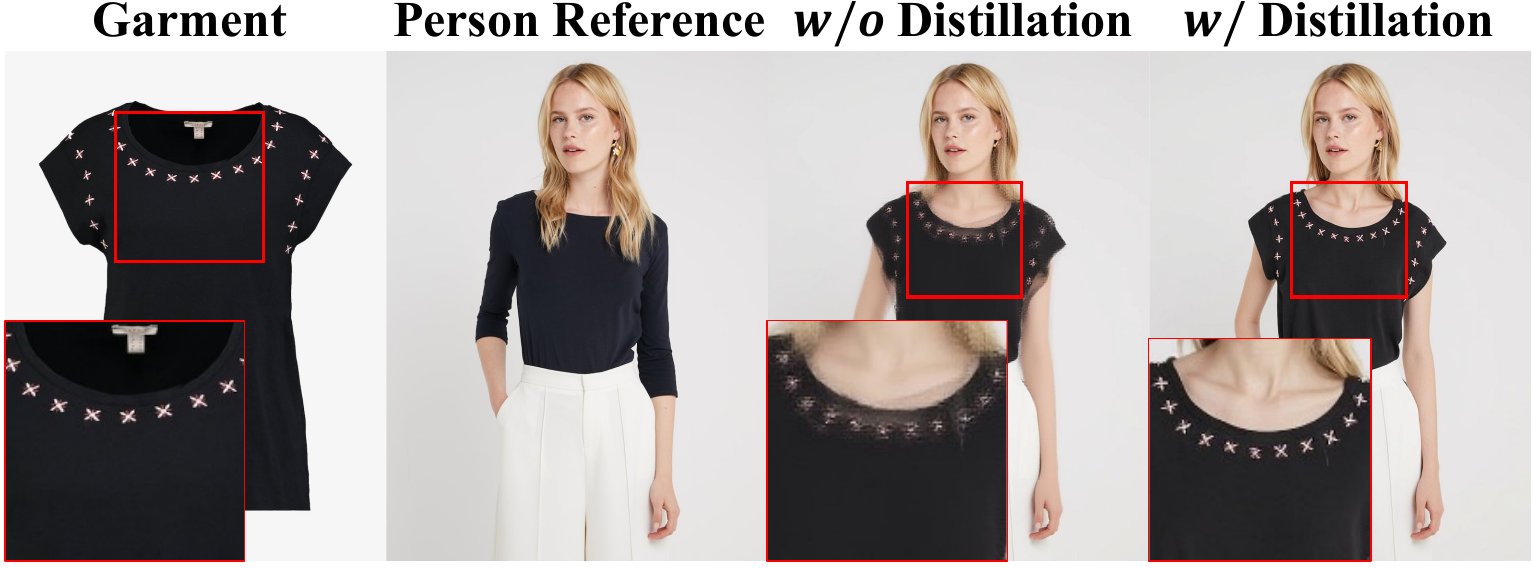} 
    \caption{Ablation study on distillation diffusion. MC-VTON without distillation diffusion brings blur and broken garment texture, decreasing the visual quality.}
    \label{ab2}
\end{figure}

We disable the Distill-LoRAs to test MC-VTON without distillation diffusion. Since the official inference step of FLUX.1-dev is set to 50, directly generating with 8 steps brings blur and broken garment texture, which decreases the visual quality, as shown in Fig.~\ref{ab2}. With distillation diffusion, MC-VTON generates realistic try-on images within 8 steps. Tab.~\ref{ab3} also demonstrates significant improvement for 8-step generation.

\begin{table}[t!]
\centering
\resizebox{0.9\linewidth}{!}{
\begin{tabular}{lccccc}
\toprule
 & LPIPS↓ & SSIM↑ &  FID↓ & KID↓ & CLIP-I↑ \\

\midrule
$w/o $ Distillation & 0.101 & 0.853  & 9.60  &1.56 & 0.885\\
$w/ $ ~~Distillation & \textbf{0.066} & \textbf{0.902} & \textbf{5.92}  &\textbf{0.773}& \textbf{0.903}\\

\bottomrule
\end{tabular}}
\caption{Ablation study on distillation diffusion. Results show that distillation diffusion achieves significant improvement for 8-step generation.}
\label{ab3}
\end{table}

\section{Conclusion}

In this paper, we introduce a DiT-based diffusion try-on method MC-VTON, which enables DiT to integrate minimal condition inputs by utilizing its intrinsic backbone in the try-on task. Experimental results confirm that MC-VTON outperforms baseline methods, demonstrating its effectiveness with superior detail fidelity, simplified network and condition inputs, reduced training parameters and inference steps. In future work, we plan to extend MC-VTON to handle video-based virtual try-on tasks.

\bibliographystyle{named}
\bibliography{main}

\begin{thebibliography}{}

\bibitem[\protect\citeauthoryear{Bi{\'n}kowski \bgroup \em et al.\egroup
  }{2018}]{kid}
Miko{\l}aj Bi{\'n}kowski, Danica~J Sutherland, Michael Arbel, and Arthur
  Gretton.
\newblock Demystifying mmd gans.
\newblock {\em arXiv preprint arXiv:1801.01401}, 2018.

\bibitem[\protect\citeauthoryear{Choi \bgroup \em et al.\egroup
  }{2021}]{vitonhd}
Seunghwan Choi, Sunghyun Park, Minsoo Lee, and Jaegul Choo.
\newblock Viton-hd: High-resolution virtual try-on via misalignment-aware
  normalization.
\newblock In {\em Proceedings of the IEEE/CVF conference on computer vision and
  pattern recognition}, pages 14131--14140, 2021.

\bibitem[\protect\citeauthoryear{Choi \bgroup \em et al.\egroup
  }{2025}]{IDM-VTON}
Yisol Choi, Sangkyung Kwak, Kyungmin Lee, Hyungwon Choi, and Jinwoo Shin.
\newblock Improving diffusion models for authentic virtual try-on in the wild.
\newblock In {\em European Conference on Computer Vision}, pages 206--235.
  Springer, 2025.

\bibitem[\protect\citeauthoryear{Esser \bgroup \em et al.\egroup }{2024}]{sd3}
Patrick Esser, Sumith Kulal, Andreas Blattmann, Rahim Entezari, Jonas
  M{\"u}ller, Harry Saini, Yam Levi, Dominik Lorenz, Axel Sauer, Frederic
  Boesel, et~al.
\newblock Scaling rectified flow transformers for high-resolution image
  synthesis.
\newblock In {\em Forty-first International Conference on Machine Learning},
  2024.

\bibitem[\protect\citeauthoryear{Gou \bgroup \em et al.\egroup
  }{2023}]{DCI-VTON}
Junhong Gou, Siyu Sun, Jianfu Zhang, Jianlou Si, Chen Qian, and Liqing Zhang.
\newblock Taming the power of diffusion models for high-quality virtual try-on
  with appearance flow.
\newblock In {\em Proceedings of the 31st ACM International Conference on
  Multimedia}, pages 7599--7607, 2023.

\bibitem[\protect\citeauthoryear{Han \bgroup \em et al.\egroup }{2018}]{viton}
Xintong Han, Zuxuan Wu, Zhe Wu, Ruichi Yu, and Larry~S Davis.
\newblock Viton: An image-based virtual try-on network.
\newblock In {\em Proceedings of the IEEE conference on computer vision and
  pattern recognition}, pages 7543--7552, 2018.

\bibitem[\protect\citeauthoryear{Ho \bgroup \em et al.\egroup }{2020}]{ddpm}
Jonathan Ho, Ajay Jain, and Pieter Abbeel.
\newblock Denoising diffusion probabilistic models.
\newblock {\em Advances in neural information processing systems},
  33:6840--6851, 2020.

\bibitem[\protect\citeauthoryear{Ho \bgroup \em et al.\egroup }{2022}]{Cascade}
Jonathan Ho, Chitwan Saharia, William Chan, David~J Fleet, Mohammad Norouzi,
  and Tim Salimans.
\newblock Cascaded diffusion models for high fidelity image generation.
\newblock {\em The Journal of Machine Learning Research}, 23(1):2249--2281,
  2022.

\bibitem[\protect\citeauthoryear{Hu \bgroup \em et al.\egroup }{2021}]{lora}
Edward~J Hu, Phillip Wallis, Zeyuan Allen-Zhu, Yuanzhi Li, Shean Wang, Lu~Wang,
  Weizhu Chen, et~al.
\newblock Lora: Low-rank adaptation of large language models.
\newblock In {\em International Conference on Learning Representations}, 2021.

\bibitem[\protect\citeauthoryear{Jiang \bgroup \em et al.\egroup
  }{2024}]{fitdit}
Boyuan Jiang, Xiaobin Hu, Donghao Luo, Qingdong He, Chengming Xu, Jinlong Peng,
  Jiangning Zhang, Chengjie Wang, Yunsheng Wu, and Yanwei Fu.
\newblock Fitdit: Advancing the authentic garment details for high-fidelity
  virtual try-on.
\newblock {\em arXiv preprint arXiv:2411.10499}, 2024.

\bibitem[\protect\citeauthoryear{Kingma}{2014}]{adam}
Diederik~P Kingma.
\newblock Adam: A method for stochastic optimization.
\newblock {\em arXiv preprint arXiv:1412.6980}, 2014.

\bibitem[\protect\citeauthoryear{Labs}{2023}]{flux}
Black~Forest Labs.
\newblock Flux.
\newblock \url{https://github.com/black-forest-labs/flux}, 2023.

\bibitem[\protect\citeauthoryear{Liu \bgroup \em et al.\egroup
  }{}]{rectifiedflow}
Xingchao Liu, Chengyue Gong, et~al.
\newblock Flow straight and fast: Learning to generate and transfer data with
  rectified flow.
\newblock In {\em The Eleventh International Conference on Learning
  Representations}.

\bibitem[\protect\citeauthoryear{Mishchenko and Defazio}{2024}]{prodigy}
Konstantin Mishchenko and Aaron Defazio.
\newblock Prodigy: An expeditiously adaptive parameter-free learner.
\newblock In {\em Forty-first International Conference on Machine Learning},
  2024.

\bibitem[\protect\citeauthoryear{Morelli \bgroup \em et al.\egroup
  }{2022}]{dresscode}
Davide Morelli, Matteo Fincato, Marcella Cornia, Federico Landi, Fabio Cesari,
  and Rita Cucchiara.
\newblock {Dress Code: High-Resolution Multi-Category Virtual Try-On}.
\newblock In {\em Proceedings of the European Conference on Computer Vision},
  2022.

\bibitem[\protect\citeauthoryear{Morelli \bgroup \em et al.\egroup
  }{2023}]{ladi-vton}
Davide Morelli, Alberto Baldrati, Giuseppe Cartella, Marcella Cornia, Marco
  Bertini, and Rita Cucchiara.
\newblock Ladi-vton: Latent diffusion textual-inversion enhanced virtual
  try-on.
\newblock In {\em Proceedings of the 31st ACM International Conference on
  Multimedia}, pages 8580--8589, 2023.

\bibitem[\protect\citeauthoryear{Mou \bgroup \em et al.\egroup }{2023}]{T2I}
Chong Mou, Xintao Wang, Liangbin Xie, Yanze Wu, Jian Zhang, Zhongang Qi, Ying
  Shan, and Xiaohu Qie.
\newblock T2i-adapter: Learning adapters to dig out more controllable ability
  for text-to-image diffusion models, 2023.

\bibitem[\protect\citeauthoryear{Nichol \bgroup \em et al.\egroup
  }{2022}]{GLIDE}
Alexander~Quinn Nichol, Prafulla Dhariwal, Aditya Ramesh, Pranav Shyam, Pamela
  Mishkin, Bob Mcgrew, Ilya Sutskever, and Mark Chen.
\newblock {GLIDE}: Towards photorealistic image generation and editing with
  text-guided diffusion models.
\newblock In Kamalika Chaudhuri, Stefanie Jegelka, Le~Song, Csaba Szepesvari,
  Gang Niu, and Sivan Sabato, editors, {\em Proceedings of the 39th
  International Conference on Machine Learning}, volume 162 of {\em Proceedings
  of Machine Learning Research}, pages 16784--16804. PMLR, 17--23 Jul 2022.

\bibitem[\protect\citeauthoryear{Peebles and Xie}{2023}]{dit}
William Peebles and Saining Xie.
\newblock Scalable diffusion models with transformers.
\newblock In {\em Proceedings of the IEEE/CVF International Conference on
  Computer Vision}, pages 4195--4205, 2023.

\bibitem[\protect\citeauthoryear{Radford \bgroup \em et al.\egroup
  }{2021}]{CLIP}
Alec Radford, Jong~Wook Kim, Chris Hallacy, Aditya Ramesh, Gabriel Goh,
  Sandhini Agarwal, Girish Sastry, Amanda Askell, Pamela Mishkin, Jack Clark,
  et~al.
\newblock Learning transferable visual models from natural language
  supervision.
\newblock In {\em International conference on machine learning}, pages
  8748--8763. PMLR, 2021.

\bibitem[\protect\citeauthoryear{Rombach \bgroup \em et al.\egroup
  }{2022}]{LDM}
Robin Rombach, Andreas Blattmann, Dominik Lorenz, Patrick Esser, and Bj{\"o}rn
  Ommer.
\newblock High-resolution image synthesis with latent diffusion models.
\newblock In {\em Proceedings of the IEEE/CVF conference on computer vision and
  pattern recognition}, pages 10684--10695, 2022.

\bibitem[\protect\citeauthoryear{Sauer \bgroup \em et al.\egroup
  }{2024}]{sd3turbo}
Axel Sauer, Frederic Boesel, Tim Dockhorn, Andreas Blattmann, Patrick Esser,
  and Robin Rombach.
\newblock Fast high-resolution image synthesis with latent adversarial
  diffusion distillation.
\newblock In {\em SIGGRAPH Asia 2024 Conference Papers}, pages 1--11, 2024.

\bibitem[\protect\citeauthoryear{Sauer \bgroup \em et al.\egroup }{2025}]{add}
Axel Sauer, Dominik Lorenz, Andreas Blattmann, and Robin Rombach.
\newblock Adversarial diffusion distillation.
\newblock In {\em European Conference on Computer Vision}, pages 87--103.
  Springer, 2025.

\bibitem[\protect\citeauthoryear{Song \bgroup \em et al.\egroup }{2020}]{ddim}
Jiaming Song, Chenlin Meng, and Stefano Ermon.
\newblock Denoising diffusion implicit models.
\newblock In {\em International Conference on Learning Representations}, 2020.

\bibitem[\protect\citeauthoryear{Tan \bgroup \em et al.\egroup
  }{2024}]{ominicontrol}
Zhenxiong Tan, Songhua Liu, Xingyi Yang, Qiaochu Xue, and Xinchao Wang.
\newblock Ominicontrol: Minimal and universal control for diffusion
  transformer.
\newblock {\em arXiv preprint arXiv:2411.15098}, 2024.

\bibitem[\protect\citeauthoryear{Wang \bgroup \em et al.\egroup
  }{2018}]{cpvton}
Bochao Wang, Huabin Zheng, Xiaodan Liang, Yimin Chen, Liang Lin, and Meng Yang.
\newblock Toward characteristic-preserving image-based virtual try-on network.
\newblock In {\em Proceedings of the European conference on computer vision
  (ECCV)}, pages 589--604, 2018.

\bibitem[\protect\citeauthoryear{Xie \bgroup \em et al.\egroup }{2023}]{gpvton}
Zhenyu Xie, Zaiyu Huang, Xin Dong, Fuwei Zhao, Haoye Dong, Xijin Zhang, Feida
  Zhu, and Xiaodan Liang.
\newblock Gp-vton: Towards general purpose virtual try-on via collaborative
  local-flow global-parsing learning.
\newblock In {\em Proceedings of the IEEE/CVF Conference on Computer Vision and
  Pattern Recognition}, pages 23550--23559, 2023.

\bibitem[\protect\citeauthoryear{Ye \bgroup \em et al.\egroup
  }{2023}]{ip-adapter}
Hu~Ye, Jun Zhang, Sibo Liu, Xiao Han, and Wei Yang.
\newblock Ip-adapter: Text compatible image prompt adapter for text-to-image
  diffusion models.
\newblock {\em arXiv preprint arXiv:2308.06721}, 2023.

\bibitem[\protect\citeauthoryear{Zhang and Agrawala}{2023}]{controlnet}
Lvmin Zhang and Maneesh Agrawala.
\newblock Adding conditional control to text-to-image diffusion models.
\newblock {\em arXiv preprint arXiv:2302.05543}, 2023.

\end{thebibliography}

\end{document}